\documentclass{article}
\usepackage{spconf,amsmath,graphicx,hyperref,amssymb}

\newcommand*{\email}[1]{%
    \normalsize\href{mailto:#1}{\texttt{#1}}\par
    }
\title{A ROBUST PIPELINE FOR CLASSIFICATION AND DETECTION OF BLEEDING FRAMES IN WIRELESS CAPSULE ENDOSCOPY USING SWIN TRANSFORMER AND RT-DETR}
\name{Sasidhar Alavala\thanks{The Auto-WCEBleedGen Challenge V1 2023 is part of the 8\textsuperscript{th} International Conference on Computer Vision \& Image Processing (CVIP-2023). Our solution stood first in that challenge.}, Anil Kumar Vadde, Aparnamala Kancheti, Subrahmanyam Gorthi}
\address{Department of Electrical Engineering, IIT Tirupati, India \\ \email{ansr2510@gmail.com}, \email{\{ee21b050, ee21b023, s.gorthi\}@iittp.ac.in}}
\begin{document}
\ninept
\maketitle
\begin{abstract}
In this paper, we present our approach to the Auto WCEBleedGen Challenge V2 2024. Our solution combines the Swin Transformer for the initial classification of bleeding frames and RT-DETR for further detection of bleeding in Wireless Capsule Endoscopy (WCE), enhanced by a series of image preprocessing steps. These steps include converting images to Lab colour space, applying Contrast Limited Adaptive Histogram Equalization (CLAHE) for better contrast, and using Gaussian blur to suppress artefacts. The Swin Transformer utilizes a tiered architecture with shifted windows to efficiently manage self-attention calculations, focusing on local windows while enabling cross-window interactions. RT-DETR features an efficient hybrid encoder for fast processing of multi-scale features and an uncertainty-minimal query selection for enhanced accuracy. The class activation maps by Ablation-CAM are plausible to the model's decisions. On the validation set, this approach achieves a classification accuracy of 98.5\% (best among the other state-of-the-art models) compared to 91.7\% without any pre-processing and an AP\textsubscript{50} of 66.7\% compared to 65.0\% with state-of-the-art YOLOv8. On the test set, this approach achieves a classification accuracy and F1 score of 87.0\% and 89.0\% respectively.
\end{abstract}

\section{INTRODUCTION}
\label{sec:introduction}
The large volume of data generated in a WCE video and the nuances of the pathological features, such as bleeding, pose considerable challenges for manual analysis by a gastroenterologist. Therefore there is a need for automated, accurate, and efficient diagnostic tools \cite{bernal2021computer}. The task in this challenge \cite{hub2024auto} is to develop an automatic pipeline which can classify each frame in a WCE video into bleeding or non-bleeding and then subsequently detect and segment the bleeding regions in that frame. Our approach to this challenge focused on two aspects:
\begin{enumerate}
\item Data pre-processing and augmentation: To overcome some problems due to artefacts in WCE frames and also limited and repetitive images, we used some efficient preprocessing and different augmentation techniques.
\item Pipeline design: To be able to accurately classify, detect and segment the bleeding frames we have used SOTA models SwinV2 \cite{liu2021swin,liu2022swin}, RT-DETR \cite{lv2023detrs} and SwinUNETR \cite{hatamizadeh2021swin}. To the best of our knowledge, this seems to be the first work to explore the usage of RT-DETR in medical imaging. We also incorporate Ablation-CAM \cite{ramaswamy2020ablation} to provide visual explanations for the classification model.
\end{enumerate}
According to the official evaluation of the challenge, our proposed pipeline achieved an accuracy of 0.88, an AP\textsubscript{50} of 70.0 and a dice score of 0.58 on test-2 data. The results show that our pipeline is robust and accurate in classifying and detecting bleeding frames and stands third among the participants.

\section{METHODOLOGY}
\label{sec:method}

\subsection{Data Preprocessing and Augmentation}
\label{ssec:subhead}
These steps include converting RGB images to Lab colour space, applying Contrast Limited Adaptive Histogram Equalization (CLAHE) for better contrast and minimising the artefacts due to bubbles, and using Gaussian blur to suppress the sharp features in G and B channels to mitigate the issues associated with high values of component a that does not correspond to the bleeding in the frame \cite{bernal2021computer}. We add several data augmentations to ensure the model's robustness since the data is limited. These include random horizontal and vertical flips, random rotations, gaussian blurring, random affine transformations, random perspective distortions, and MixUp as these augmentations.

\subsection{Pipeline}
\label{ssec:subhead}

\begin{figure}[htb]

\begin{minipage}[b]{1.0\linewidth}
  \centering
  \centerline{\includegraphics[width=8.5cm]{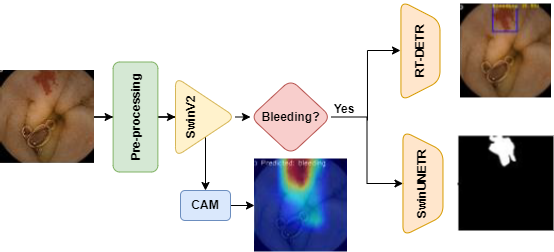}}
  \vspace{0cm}
\end{minipage}
\caption{Proposed bleeding detection pipeline.}
\label{fig:fig1}
\end{figure}

\begin{figure}[htb]
\begin{minipage}[b]{1.0\linewidth}
  \centering
  \centerline{\includegraphics[width=8.5cm]{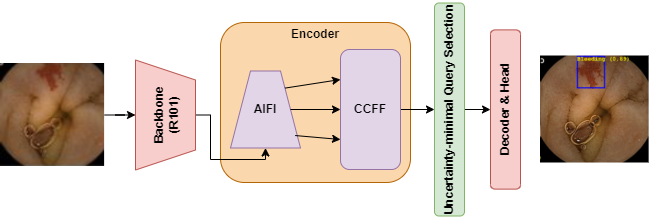}}
  \vspace{0cm}
\end{minipage}
\caption{Block level RT-DETR architecture \cite{lv2023detrs}.}
\label{fig:fig2}
\end{figure}
Our approach includes initially classifying the frame into bleeding or non-bleeding and then detecting and segmenting the bleeding regions in that frame.

SwinV2 and RT-DETR are employed for their robust performance in detecting and classifying intricate details within images. Ablation-CAM \cite{ramaswamy2020ablation} in this pipeline introduces a layer of interpretability, offering visual explanations behind the classification decisions made by the model. While SwinUNETR \cite{hatamizadeh2021swin} is used to segment the bleeding frames, it combines patch-based processing and transformer encoders with a U-Net style decoder

Real-Time Detection Transformer, RT-DETR \cite{lv2023detrs}, the current state-of-the-art end-to-end detection model, incorporates an efficient hybrid encoder that processes multi-scale features to speed up inference and introduces uncertainty-minimal query selection to improve initial query quality for the decoder, boosting overall accuracy. Attention-based Intra-scale Feature Interaction (AIFI) and the CNN-based Cross-scale Feature Fusion (CCFF) make up the encoder block optimizing the processing of multi-scale features.

\subsection{Training Details}
\label{ssec:subhead}
The input image size is 224x224 pixels trained with a single NVIDIA RTXA4000.
\begin{enumerate}
\item For SwinV2-T: Model is trained using Cross-Entropy loss and the AdamW optimizer with an initial learning rate of 0.0001, adjusted by a Cosine Annealing scheduler.
\item For RT-DETR: Model is trained using a combination of focal loss and a mix of L1 and generalized IoU losses for bounding box regression. The uncertainty in the object queries is also integrated into the loss function \cite{lv2023detrs}. The training employs the AdamW optimizer with a specific learning rate strategy that includes a warm-up and subsequent decay phase.
\item For SwinUNETR: Model is trained using a combination of Cross-Entropy, Dice losses (with a weightage of 0.3 and 0.7 respectively) and the Adam optimizer with a learning rate of 0.0001.
\end{enumerate}

\section{RESULTS}
\label{sec:results}

\begin{figure}[htb]
\begin{minipage}[b]{1.0\linewidth}
  \centering
  \centerline{\includegraphics[width=8.5cm]{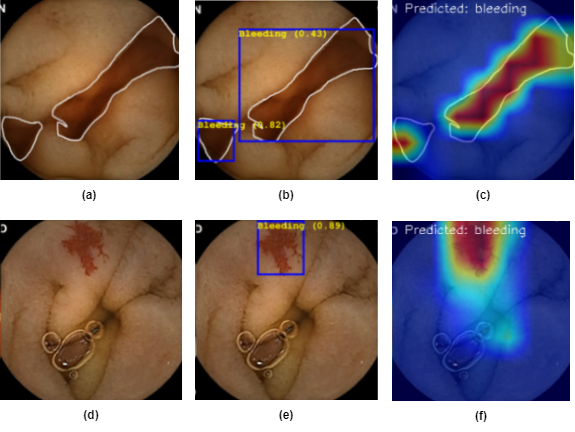}}
  \vspace{0cm}
\end{minipage}
\caption{Visual results of our approach on test-1 marked images (top row) and test-2 unmarked images (bottom row).}
\label{fig:fig3}
\end{figure}

Figure 3 shows the accuracy of the classification and detection model in predicting the bleeding regions and class activation maps also support our visual reasoning. Table 1 presents a comparison of classification results on the validation set, highlighting the impact of our preprocessing on model performance. The results are listed for several models, including EfficientNet-B5, ViT-B, DeIT-S, Swin-T, and SwinV2-T, with a clear indication of accuracy improvements. SwinV2-T model outperforms the others, achieving a remarkable accuracy of 98.5\% with our preprocessing, compared to its initial 91.7\% accuracy. 

Table 2 presents a comparison of detection results on the validation set for some methods, comparing them based on the Average Precision (AP) metric at different Intersections over Union (IoU) thresholds. RT-DETR-R101 model demonstrates superior performance with an AP\textsubscript{50} of 66.7\% and an AP\textsubscript{50:95} of 81.0\%, outpacing the state-of-the-art YOLOv8-L. RT-DETR model not only has detection accuracy but also better convergence compared to the traditional DETR model. For segmentation, SwinUNETR yielded a dice score of 0.62 and an IoU of 0.48 on the validation set.

\begin{table}[h]
\centering
\caption{Classification results on validation set showing accuracy with and without our pre-processing.}
\label{tab:table1}
\setlength{\tabcolsep}{7pt} 
\renewcommand{\arraystretch}{1.3} 
\begin{tabular}{|c|c|p{1.5cm}|p{1.5cm}|}
\hline
\textbf{Method} & \textbf{\#param.} & \textbf{Accuracy\%} \\
\hline
EfficientNet-B5 \cite{tan2019efficientnet} & 25M & 85.2/95.3 \\
\hline
ViT-B \cite{dosovitskiy2020image} & 40M & 86.1/96.5 \\
\hline
DeIT-S \cite{touvron2021training} & 20M & 86.2/96.9 \\
\hline
Swin-T \cite{liu2021swin} & 27M & 88.7/97.8 \\
\hline
\textbf{SwinV2-T} \cite{liu2022swin} & \textbf{37M} & \textbf{91.7/98.5} \\
\hline
\end{tabular}
\end{table}

\begin{table}[ht!]
\centering
\caption{Detection results on validation set.}
\label{tab:table2}
\setlength{\tabcolsep}{7pt} 
\renewcommand{\arraystretch}{1.3} 
\begin{tabular}{|c|c|c|c|c|}
\hline
\textbf{Method} & \textbf{\#param.} & \textbf{Epochs} & \textbf{AP\textsubscript{50}/AP\textsubscript{50:95}} \\
\hline
YOLOv5-L \cite{yolov5} & 45M & 150 & 63.2/75.6 \\
\hline
YOLOv8-L \cite{yolov8_ultralytics} & 43M & 150 & 65.0/79.7 \\
\hline
DETR-DC5-R101 \cite{carion2020end} & 58M & 500 & 61.2/72.3 \\
\hline
\textbf{RT-DETR-R101} \cite{lv2023detrs} & \textbf{75M} & \textbf{150} & \textbf{66.7/81.0} \\
\hline
\end{tabular}
\end{table}

\section{CONCLUSION}
\label{sec:conclusion}
 In conclusion, our approach to this Auto-BleedGEN V2 2024 challenge, has been able to accurately classify, detect and segment the bleeding regions in WCE frames. The class activation maps also corroborate the model's decision-making process, aligning closely with the identified areas of interest. Future work includes addressing the cases where the model failed (including when blood vessels are prominent, small bleeding regions, and illumination conditions) and also addressing the multiple false positives in detection.

\vfill\pagebreak

\section{ACKNOWLEDGEMENT}
\label{sec:ack}
This research has received partial support through project grant number 2022-IRP-26694272 from the Semiconductor Research Corporation (SRC) under the India Research Program (IRP).

\bibliographystyle{IEEEbib}
\bibliography{refs}

\end{document}